\documentclass[conference]{IEEEtran}
\IEEEoverridecommandlockouts
\usepackage{cite}
\usepackage{amsmath,amssymb,amsfonts}
\usepackage[ruled,vlined]{algorithm2e}
\usepackage{float}
\SetKwInput{KwIn}{Input}
\SetKwInput{KwOut}{Output}
\usepackage{graphicx}
\usepackage{textcomp}
\usepackage{url} 
\usepackage{float}
\usepackage{booktabs}
\usepackage{xcolor}
\usepackage{subfig}

\def\BibTeX{{\rm B\kern-.05em{\sc i\kern-.025em b}\kern-.08em
    T\kern-.1667em\lower.7ex\hbox{E}\kern-.125emX}}
\begin{document}

\title{EARL: Energy-Aware Optimization of Liquid State Machines for Pervasive AI}

\author{\IEEEauthorblockN{ Zain Iqbal}
\IEEEauthorblockA{\textit{Institute for Informatics and Telematics
} \\
\textit{National Research Council}\\
Pisa, Italy \\
zain.iqbal@iit.cnr.it}
\and
\IEEEauthorblockN{ Lorenzo Valerio}
\IEEEauthorblockA{\textit{Institute for Informatics and Telematics
} \\
\textit{National Research Council}\\
Pisa, Italy \\
lorenzo.valerio@iit.cnr.it}
}
\maketitle

\begin{abstract}
Pervasive AI increasingly depends on on-device learning systems that deliver low-latency, energy-efficient computation under strict resource constraints. Liquid State Machines (LSMs) offer a promising path for low-power temporal processing in pervasive and neuromorphic systems, but their deployment remains difficult due to high hyperparameter sensitivity and the computational cost of traditional optimization methods that ignore energy constraints. This work presents EARL, an energy-aware reinforcement learning framework that integrates Bayesian optimization with an adaptive RL-based selection policy to jointly optimize accuracy and energy use. EARL employs surrogate modeling for global exploration, reinforcement learning for dynamic candidate prioritization, and an early-termination mechanism to remove redundant evaluations, substantially reducing computational overhead. Experiments on three benchmark datasets show that EARL achieves 6–15\% higher accuracy, 60–80\% lower energy consumption, and up to an order-of-magnitude reduction in optimization time compared to leading hyperparameter tuning frameworks. These results highlight the effectiveness of energy-aware adaptive search in improving the efficiency and scalability of LSMs for resource-constrained on-device AI applications.
\end{abstract}


\section{Introduction}

The rapid expansion of machine learning into edge, embedded, and neuromorphic platforms has increased the demand for models that deliver high predictive performance under strict computational and energy constraints. 
While modern deep architectures, e.g., attention-based networks, Vision Transformers, and LLMs, continue to advance state-of-the-art accuracy, they demand substantial compute and memory resources \cite{strubell2019energy}, restricting their deployment in latency- and power-limited environments. Beyond inference, emerging pervasive AI applications increasingly require on-device training, motivated by privacy requirements, sensitivity of personal data, limited communication bandwidth, and the need for continual learning from locally generated sensor streams. These constraints have renewed interest in brain-inspired and energy-efficient computing paradigms that inherently support adaptive, low-power computation.

Reservoir Computing (RC) \cite{Maass2002} has emerged as an attractive alternative due to its architectural simplicity, low training cost, and suitability for real time processing. Within RC, the Liquid State Machine (LSM) leverages sparse spiking neuron dynamics and event driven computation, making it particularly relevant for neuromorphic hardware \cite{gallicchio2011echo}. Despite these advantages, LSM performance is highly sensitive to hyperparameters such as spectral radius, leak rate, reservoir size, and synaptic connectivity. Small deviations in these parameters can shift the reservoir between stable, edge of chaos, and chaotic regimes, causing significant variation in both accuracy and energy consumption \cite{Bianchi2020, gallicchio2011echo}. As a result, principled hyperparameter optimization (HPO) is essential for reliable deployment.

Conventional HPO methods—including grid search, random search \cite{Bergstra2012}, and Bayesian optimization \cite{Snoek2012}—perform poorly for spiking reservoirs because they assume smooth, continuous objective landscapes and do not treat energy consumption as an explicit optimization target. These assumptions break down in the discrete, nonlinear, and sometimes chaotic dynamics of LSMs, where evaluating each configuration is costly and the search space contains many local optima.



In this work, we introduce the Energy Aware Reinforcement Learning (EARL) framework, a hybrid optimization strategy designed specifically for LSM hyperparameters. EARL combines Bayesian Optimization for global exploration with a reinforcement learning agent that adaptively selects candidates according to a reward function that jointly encodes a performance measure, e.g., classification accuracy, and reservoir level energy usage. An early termination mechanism halts the search when additional evaluations offer negligible improvement, reducing computational overhead and avoiding unnecessary energy expenditure.

We evaluated EARL on three standard datasets. These tasks cover speech recognition, activity recognition, and environmental monitoring modalities. Across all experiments, EARL consistently improves accuracy, energy trade-offs, and reduces total optimization time compared to widely used frameworks such as Optuna \cite{akiba2019optuna} and Ray Tune \cite{moritz2018ray}.

The primary contributions of this work are:
\begin{itemize}
    \item We formulate an energy aware HPO problem for LSMs that explicitly incorporates accuracy and reservoir energy consumption into a unified reward function.
    \item We propose the EARL framework, which integrates Bayesian Optimization for global surrogate modeling with reinforcement learning for adaptive candidate selection and automatic early termination.
   \item We introduce an adaptive early-termination mechanism within EARL that halts the optimization when reward or energy improvements plateau.
\end{itemize}
The paper is organized as follows: Section~II reviews background and LSM fundamentals; Section~III introduces the proposed EARL framework; Section~IV details the experimental setup; Section~V presents results and analysis; and Section~VI concludes with future directions.

\section{Background And Related Work}
Reservoir Computing offers an efficient framework for processing temporal and sequential data, resulting in significantly lower training costs and reduced energy consumption compared to traditional deep neural networks. In RC, a fixed recurrent reservoir transforms time-varying inputs into high-dimensional internal representations, which are subsequently processed by a trainable readout layer. Among RC models, the LSM and Echo State Network (ESN) are the most prominent. Both utilize recurrent dynamics to capture temporal dependencies; however, the LSM employs spiking neurons, enabling event-driven computation and inherently low-power operation. The behavior of LSM closely mirrors the synaptic plasticity found in biological neurons, where spike timing governs learning and memory formation \cite{Schrauwen2008}, highlighting its relevance for bio-inspired and neuromorphic computing\footnote{The architecture of LSM's naturally maps onto neuromorphic hardware platforms, where asynchronous spike-based processing and local synaptic updates can be efficiently realized in emerging substrates such as memristive arrays and neuromorphic silicon.}.

\subsection{Liquid State Machine Architecture}

\emph{Liquid State Machine} are spiking neural network that processes temporal data through its intrinsic recurrent dynamics. The LSM operates in three main stages: input encoding, dynamic processing within the reservoir, and readout. The input layer converts continuous data into spike trains, which are then propagated through a randomly connected reservoir of \textit{Leaky Integrate-and-Fire} (LIF) neurons. The reservoir’s synaptic weights remain fixed during training, allowing it to transform temporal inputs into high-dimensional representations without the need for backpropagation \cite{Gallicchio2017}.

The membrane potential of each neuron evolves according to the leaky integration rule:
\begin{equation}
\small
\tau_m \frac{dV_i}{dt} = -(V_i - V_{\text{rest}}) + I_i(t),
\end{equation}
where $V_i$ denotes the membrane potential, $V_{\text{rest}}$ is the resting potential, and $I_i(t)$ represents the synaptic input current. When $V_i$ crosses a firing threshold, the neuron emits a spike and resets its potential. This mechanism provides a fading memory of previous activity, allowing the LSM to capture temporal dependencies across multiple timescales.

The reservoir state is updated as:
\begin{equation}
\small
x_t = (1 - \alpha)x_{t-1} + \alpha \tanh(Wx_{t-1} + W_{\text{in}}u_t),
\end{equation}
where $u_t$ is the input vector, $W_{\text{in}}$ and $W$ are the input and recurrent weight matrices\footnote{$W$ is scaled by its spectral radius to ensure stable dynamics while preserving the reservoir’s nonlinear richness.}, and $\alpha$ is the leak rate governing the balance between memory retention and responsiveness to new input, ranging from overly stable dynamics (e.g., $\alpha <0.1$) to chaotic behavior that degrades temporal coherence (e.g., $\alpha > 0.3$).\cite{Maass2002}. 
This configuration allows the LSM to project input sequences into a rich high-dimensional space while keeping energy consumption low. The number of neurons, connection probability, and spectral radius jointly shape the trade-off between computational cost, accuracy, and energy efficiency. We optimize these parameters later through the proposed adaptive learning framework.\


\subsection{GRU Readout and Temporal Abstraction}

To enhance temporal abstraction, the reservoir outputs are passed to \textit{Gated Recurrent Unit} (GRU) that, in this paper, serves as the trainable readout. The GRU maintains a hidden state $h_t$, updated according to:
\begin{equation}
\small
h_t = z_t \odot h_{t-1} + (1 - z_t) \odot \tilde{h}_t,
\end{equation}
where $z_t$ denotes the update gate, $\tilde{h}_t$ represents the candidate activation, and $\odot$ indicates element-wise multiplication \cite{Cho2014}. The gating mechanisms regulate how past information is retained or replaced, allowing the GRU to function as a learned temporal filter over the spiking activity of the reservoir.

This hybrid LSM-GRU configuration integrates the nonlinear, event-driven dynamics of the LSM with the adaptive memory of the GRU. During training, only the GRU and output layer parameters are updated, while the reservoir connections remain fixed. This approach preserves the computational efficiency of reservoir computing while enhancing its representational capacity for sequential and time-dependent tasks.

\subsection{Energy-Aware Optimization and Limitations of Prior Work}

RC offers strong computational efficiency, yet its performance depends heavily on hyperparameters such as spectral radius, leak rate, input scaling, and reservoir size. Small perturbations in these parameters can shift the reservoir dynamics between stable and chaotic regimes, making principled hyperparameter optimization essential \cite{Bianchi2020, Gallicchio2017}.
Early work in RC relied mainly on extensive grid search and random sampling to identify suitable configurations \cite{Bergstra2012}. Although effective in low-dimensional settings, these methods scale poorly as the search space grows. More advanced approaches \textit{including Bayesian Optimization} \cite{Snoek2012} and \textit{Hyperband-based} strategies \cite{Li2018} improved efficiency, but typically assumed smooth, continuous objective landscapes that do not reflect the nonlinear, discrete, and sometimes chaotic behavior of recurrent spiking reservoirs.
More recently, Mendula \textit{et al.} (2025) introduced an adaptive $\varepsilon$-greedy hyperparameter search strategy tailored for reservoir computing. Their framework incorporates a memory assisted exploration mechanism that dynamically balances exploration and exploitation, preserves a buffer of high performing configurations, and penalizes overly large reservoirs to reduce computational and energy costs \cite{Mendula2025}. 
However, the method was designed and evaluated exclusively on non-spiking ESN. 
%
Other studies explore intrinsic plasticity to regulate neuronal excitability \cite{Schrauwen2008} and apply reinforcement learning or meta-learning for online adjustment \cite{Stanley2019}. Although these techniques improve robustness and adaptability, they generally do not include explicit energy objectives in the optimization process.
Recent studies have also examined the integration of spiking reservoirs with recurrent readouts to improve temporal abstraction and robustness \cite{Gallicchio2017, Bianchi2020}, but most approaches still rely on offline, static hyperparameter tuning and provide only indirect estimates of energy usage.

Overall, prior research has not addressed the need for adaptive, energy-aware optimization frameworks that jointly balance accuracy and power consumption during operation. In contrast, our proposed approach introduces adaptive mechanisms that enable continuous tuning in response to changing system dynamics as described in Section \ref{PA}

\section{Proposed Approach}
\label{PA}
To optimize the hyperparameter search for LSM, we propose a hybrid  Energy-Aware Reinforcement Learning (EARL) Framework that integrates Bayesian optimization with reinforcement learning. Bayesian optimization is used to model the objective landscape using a surrogate function, allowing the selection of promising configurations based on previous observations. Reinforcement learning is introduced as a local decision-making agent that selects the next trial among candidate suggestions by maximizing a reward function that balances classification accuracy and energy consumption.

\subsection{Sobol initialization }

The LSM's performance is highly sensitive to parameters such as leak rate $\mathcal{X}_{\beta}$, spectral radius $\mathcal{X}_{\text{spectral}}$, reservoir size $\mathcal{X}_{\text{size}}$, and synaptic connectivity $\mathcal{X}_{\text{conn}}$, which form our search space: 

\begin{equation}
\small
    \mathcal{X} = 
    \mathcal{X}_{\text{size}} 
    \times 
    \mathcal{X}_{\text{conn}} 
    \times 
    \mathcal{X}_{\text{spectral}} 
    \times 
    \mathcal{X}_{\beta}.
\end{equation}

The first \( N_{\text{init}}\) hyperparameter configurations are generated using Sobol sequences\footnote{Sobol sampling provides a low-discrepancy quasi-random sequence, ensuring points are evenly distributed across \( \mathcal{X} \) without clustering or bias.}.  
Unlike pure random sampling, Sobol sequences ensure each subregion receives proportional representation, which improves the Gaussian Process model's ability to approximate the objective landscape (i.e., next phase presented in Sec. \ref{sec:BO}.

For each sampled configuration \( x_i \in \mathcal{X} \), LSM is trained and evaluated, yielding the triplet of metrics: learning performance that in our case corresponds to classification accuracy \( f_1(x_i) \), energy consumption \( f_2(x_i) \), and the reward function \( r(x_i) \) that combines both objectives:
\begin{equation} 
\small
\label{eq4}
r(x) = f_1(x) - \alpha \cdot f_2(x), 
\end{equation} 
where $\alpha$ balances the energy contribution to the reward. These evaluations are accumulated into the dataset:
\begin{equation}
\small
\mathcal{D}_t = \{(x_i, f_1(x_i), f_2(x_i), r(x_i))\}_{i=1}^{t}.
\end{equation}

During initialization, the framework tracks the best observed configuration by maintaining:
\begin{equation}
\small
r^\ast = \max_{i=1}^{N_{\text{init}}} r(x_i), \quad f_2^\ast = \min_{i=1}^{N_{\text{init}}} f_2(x_i).
\end{equation}
This baseline serves as a reference point for measuring improvement in subsequent phases.

\subsection{Bayesian Optimization with Gaussian Process Modeling}

Following the Sobol initialization phase, Bayesian Optimization (BO) transitions to a model-based regime that serves as the global controller for hyperparameter selection. This phase leverages the diverse initial dataset to construct a probabilistic surrogate model that efficiently explores the hyperparameter space.

\subsubsection{Gaussian Process Surrogate Modeling}
\label{sec:BO}
The Bayesian optimization phase uses a Gaussian Process (GP) to approximate the reward function from previously evaluated configurations. The GP models the reward as a distribution over functions:
\begin{equation}
\small
r(x) \sim \mathcal{GP}(\mu(x), k(x, x')),
\end{equation}
where \( \mu(x) \) is the predicted reward at configuration \( x \), and \( k(x, x') \) is a kernel function that measures similarity between configurations. Using the accumulated dataset \( \mathcal{D}_t \), the GP learns how different hyperparameter combinations influence performance.

The kernel encodes assumptions about the structure of the objective landscape; we adopt a Matérn kernel because it can capture both broad trends and local variations. For any untested configuration on the LSM, the GP returns a predicted mean \( \mu(x) \) and an uncertainty estimate \( \sigma^2(x) \). This uncertainty allows the optimizer to balance exploitation (selecting configurations expected to perform well) with exploration (testing uncertain regions that may yield better solutions). After each batch of evaluations, the GP is retrained with the newly collected observations to refine its model of the reward landscape (see Eq. \ref{eq:datasetupdate}). 



\subsubsection{Expected Improvement and Candidate Generation}

Based on the GP surrogate model, the framework selects promising candidate configurations using the Expected Improvement (EI) acquisition function, which measures the expected gain over the current best rewards $r^\ast$. 

\begin{equation}
\small
\begin{split}
\text{EI}(x) &= \mathbb{E}[\max(r(x) - r^\ast, 0)] \\
&= \begin{cases}
(\mu(x) - r^\ast)\Phi(Z) + \sigma(x)\phi(Z), & 
    \text{if } \sigma(x) > 0, \\
0, & \text{if } \sigma(x) = 0,
\end{cases}
\end{split}
\end{equation}

where \(
Z = \frac{\mu(x) - r^\ast}{\sigma(x)},
\)
, $\Phi$ and $\phi$ are the CDF and PDF of the standard normal distribution. 

To support parallel evaluation, the framework selects a batch of
\( K \) candidate configurations by maximizing EI:
\begin{equation}
\small
x_{t+1}^{(j)} = \arg\max_{x \in \mathcal{X}} \Gamma(x; \mathcal{D}_t), \quad j = 1, \dots, K,
\end{equation}
where \( \Gamma(x; \mathcal{D}_t) \) is the EI computed over the best observed reward\footnote{Batch generation enables parallel evaluation of multiple LSM configurations, reducing overall optimization time while improving exploration.}. To ensure diversity, each candidate \( x_{t+1}^{(j)} \) is compared with previously evaluated points, and duplicates are perturbed to avoid reduntant trials.

\subsubsection{Reward Approximation}

While optimization is guided by the scalarized reward function, the framework retains full metric evaluations $(f_1(x), f_2(x))$ for all configurations. Each candidate configuration \( x_{t+1}^{(j)} \) selected by the acquisition function is represented through its GP posterior:

\begin{equation}
\small
\label{eq11}
r(x_{t+1}^{(j)}) \sim \mathcal{N}(\mu_j, \sigma_j^2),
\end{equation}
where $\mu_j$ and $\sigma_j^2$ represent the predictive mean and variance for that candidate. This distribution captures both expected reward and uncertainty, which the reinforcement learning stage subsequently exploits.

Following each batch evaluation, the dataset is updated as:
\begin{equation}
\small
\label{eq:datasetupdate}
\mathcal{D}_{t+1} = \mathcal{D}_t \cup \{(x_{t+1}^{(j)}, f_1(x_{t+1}^{(j)}),f_2(x_{t+1}^{(j)}), r(x_{t+1}^{(j)}))\}.
\end{equation}

After optimization, the complete set of evaluations is used to approximate the Pareto frontier:
\begin{equation}
\begin{split}
\small
\label{eq10}
\mathcal{P}^* = 
\Bigl\{\, x \in X \;\Big|\; 
& \nexists x' \in X : \\
& \bigl(f_1(x') \ge f_1(x) \land f_2(x') \le f_2(x)\bigr) \\
& \land \bigl(f_1(x') > f_1(x) \lor f_2(x') < f_2(x)\bigr)
\Bigr\}.
\end{split}
\end{equation}
This frontier provides principled basis for selecting LSM configurations that jointly optimize accuracy and energy efficiency.

\subsection{Reinforcement Learning-Guided Selection}

While Bayesian Optimization (BO) provides probabilistically informed candidates, it does not reason about long-term performance trends. To address this limitation, we incorporate a Reinforcement Learning (RL) agent that selects the most promising configuration from each batch of $K$ candidates proposed by BO.

\subsubsection{State Representation and Action Selection}
For each candidate \(x_{t+1}^{(j)}\), the GP posterior \(r(x_{t+1}^{(j)}) \sim \mathcal{N}(\mu_j,\sigma_j^2)\) provides predictive statistics. The RL state is the normalized feature vector
\[
\small
s_t = [\tilde{\mu}_1,\tilde{\sigma}_1^2,\dots,\tilde{\mu}_K,\tilde{\sigma}_K^2],
\]
where min--max normalization is applied across candidates. The agent selects an action \(a_t \in \{1,\dots,K\}\) using an \(\epsilon\)-greedy rule, 
\begin{equation}
\small
a_t = 
\begin{cases}
\text{uniform}(1,\dots,K) & \text{with probability } \epsilon\\
\arg\max_j Q(s_t, j) & \text{otherwise}
\end{cases}
\end{equation}
with
\[
\small
\epsilon_t = \max(\kappa \epsilon_{t-1}, \epsilon_{\min}),
\]
to gradually shift from exploration to exploitation.



\subsubsection{Replay Buffer and Learning}
Transitions \((s_t, a_t, r_t, s_{t+1})\) are stored in a replay buffer of capacity \(C\), enabling decorrelated and repeated sampling. The buffer operates as a first-in-first-out  (FIFO) queue.

The RL agent approximates its action-value function \(Q(s,a)\) using a single neural network updated periodically rather than at every step. Every \(F\) iterations (\(t \bmod F = 0\)), a mini-batch of transitions is sampled from the replay buffer, and the network is refined via one-step temporal-difference (TD) learning.
The update rule is
\begin{equation}
\small
Q(s_t,a_t) \leftarrow Q(s_t,a_t)
+ \eta \big[r_t + \gamma \max_{a'} \bar{Q}(s_{t+1},a') - Q(s_t,a_t)\big],
\end{equation}
where \(\eta\) is the learning rate and \(\gamma\) the discount factor. The term
\[
\small
r_t + \gamma \max_{a'} \bar{Q}(s_{t+1},a')
\]
serves as the TD target, and its difference from the current estimate,
\[
\small
\delta_t = r_t + \gamma \max_{a'} \bar{Q}(s_{t+1},a') - Q(s_t,a_t),
\]
defines the TD error that drives parameter updates.

A separate target network \(\bar{Q}\) provides the bootstrap values in the TD target and is synchronized with the main network every \(F\) steps. This decoupling stabilizes training by reducing feedback loops between evolving predictions and targets.

\subsubsection{Adaptive Early Termination}

To avoid unnecessary evaluations during stagnation, the framework employs an
adaptive early termination criterion. Let \(r^\ast\) and \(f_2^\ast\) denote the
best reward and lowest energy observed so far:
\[
\small
r^\ast = \max_{i < t} r(x_i), \qquad
f_2^\ast = \min_{i < t} f_2(x_i).
\]

Relative improvements at iteration \(t\) are defined as
\[
\small
\Delta r_t = r(x_t) - r^\ast, \qquad
\Delta f_{2,t} = f_2(x_t) - f_2^\ast.
\]

The framework monitors these values over a sliding window of \(T_s\) recent evaluations. If neither objective exhibits progress beyond tolerance thresholds \((\epsilon_r, \epsilon_e)\) for \(T_s\) consecutive steps, 
\begin{equation}
\forall j \in \{t - T_s + 1, \dots, t\}:\;
\Delta r_j \le \epsilon_r \;\land\;
\Delta f_{2,j} \ge -\epsilon_e,
\end{equation}
the optimization loop is terminated. This stopping rule detects convergence to regions of marginal improvement and prevents excessive computation in performance plateaus, enhancing the overall efficiency of the EARL search process.

\section{Experimental Setup}
We evaluate the proposed \emph{EARL} framework on three public benchmark datasets covering audio, environmental sensing, and human activity recognition: the Free Spoken Digit Dataset (FSDD), Occupancy Detection, and UCI Human Activity Recognition (HAR).

The FSDD dataset \cite{fsdd} provides $\sim$3,000 spoken-digit audio samples. We resample each waveform and extract MFCC features for classification. The Occupancy Detection dataset \cite{occupancy} offers $\sim$10,000 environmental sensor measurements labeled as occupied or unoccupied, forming a representative low-dimensional binary task. The
UCI HAR dataset \cite{har} contains $\sim$15,000 multivariate accelerometer and gyroscope sequences across six activity classes and serves as a standard benchmark for time-series models.
We normalize all datasets to zero mean and unit variance and split them into 80\% training and 20\% validation using stratified sampling. Across all experiments, a Liquid State Machine (LSM) generates reservoir states, and a GRU network performs classification. We optimize LSM hyperparameters (leak rate, spectral radius, connectivity, reservoir size) as summarized in Table~\ref{tab:hyperparams}, and train the GRU for 100 epochs using AdamW. All models run in PyTorch with fixed random seeds to ensure reproducibility.

We conduct all experiments on an NVIDIA Tesla T4 GPU (16 GB VRAM) under identical runtime settings. For each trial, we record classification accuracy, energy consumption, and execution time.
To evaluate optimization performance, EARL was compared with two widely used baselines;Optuna and Ray Tune. All methods were configured with the same search space, parameter bounds, and trial budget (50 trials, including 20 initialization trials). Optuna employed the NSGA-II multi-objective sampler, whereas Ray Tune utilized asynchronous parallel sampling. All algorithms were evaluated using identical training scripts, datasets, and experimental protocols to ensure a fair comparison. Results represent the mean of 100 repeated runs to reduce stochastic variation.
\begin{table}[htbp]
\centering
\caption{Hyperparameter search space used across all experiments.\\}
\begin{tabular}{ll}
\toprule
\textbf{Hyperparameter} & \textbf{Search Space} \\
\midrule
Leak Rate          & [0.1, 0.4] (Linear-uniform) \\
Spectral Radius         & [0.6, 1.1] (Linear-uniform) \\
Reservoir Size          & [100, 1000] (Integer-uniform) \\
Connectivity               & [0.2, 0.7] (Linear-uniform) \\
Activation Function             & \{\texttt{tanh}\} (Fixed) \\
Seed                            & 42 (Fixed) \\
Total Trials      & 50 (Fixed) \\
Initial Trials      & 20 (Fixed) \\
Classifier Epochs    & 100 (Fixed) \\
Batch Size                      & 64 (Fixed) \\
\bottomrule
\end{tabular}
\label{tab:hyperparams}
\end{table}

\begin{table*}[htbp!]
\caption{Comparative performance of EARL, Optuna, and Ray across all datasets}
\label{table2}
\centering
\scriptsize
\resizebox{\textwidth}{!}{
\begin{tabular}{l l c c c c}
\hline
\textbf{Dataset} & \textbf{Model} & \textbf{Accuracy (\% $\pm$ CI)} & \textbf{Energy (pJ/sample $\pm$ CI)} & \textbf{Total Train Time (min)} & \textbf{Total Optimization Time (min)} \\
\hline
\textbf{FSDD} & \textbf{EARL}   & \textbf{95.39 $\pm$ 0.44}   & \textbf{0.2089 $\pm$ 0.0129}  & 10.31 & \textbf{10.57} \\
              & Optuna          & 82.25 $\pm$ 2.75             & 0.3644 $\pm$ 0.0091           & 86.33 & 106.60 \\
              & Ray             & 88.15 $\pm$ 2.18             & 0.4178 $\pm$ 0.0042           & 80.85 & 100.88 \\
\hline
\textbf{HAR}  & \textbf{EARL}   & \textbf{96.99 $\pm$ 0.34}    & \textbf{0.20796 $\pm$ 0.02362} & 18.73 & \textbf{19.53} \\
              & Optuna          & 90.52 $\pm$ 1.16             & 0.6265 $\pm$ 0.0193           & 38.73 & 47.75 \\
              & Ray             & 94.41 $\pm$ 0.67             & 0.43737 $\pm$ 0.00512         & 47.77 & 58.48 \\
\hline
\textbf{Occupancy} & \textbf{EARL} & \textbf{98.47 $\pm$ 0.00008} & \textbf{0.0278 $\pm$ 0.0052}  & 14.42 & \textbf{15.28} \\
                   & Optuna        & 97.44 $\pm$ 0.27             & 0.1599 $\pm$ 0.0092           & 38.73 & 47.75 \\
                   & Ray           & 98.47 $\pm$ 0.03             & 0.1624 $\pm$ 0.0030           & 52.74 & 64.85 \\
\hline
\end{tabular}
}
\label{tab:earl_comparison}
\end{table*}

\section{Results and Discussion}

The following sections present the performance trends observed across the three datasets.
Table \ref{table2} presents a quantitative comparison of EARL, Optuna, and Ray Tune across all benchmark datasets. Figures \ref{fig:accuracy}–\ref{fig:pareto} further illustrate trial-wise accuracy trajectories, energy consumption trends, and resulting Pareto fronts. 
\subsubsection{Accuracy Performance}
\begin{figure}[htbp!]
\centering
\subfloat[FSDD Dataset]{
\includegraphics[width=0.5\columnwidth]{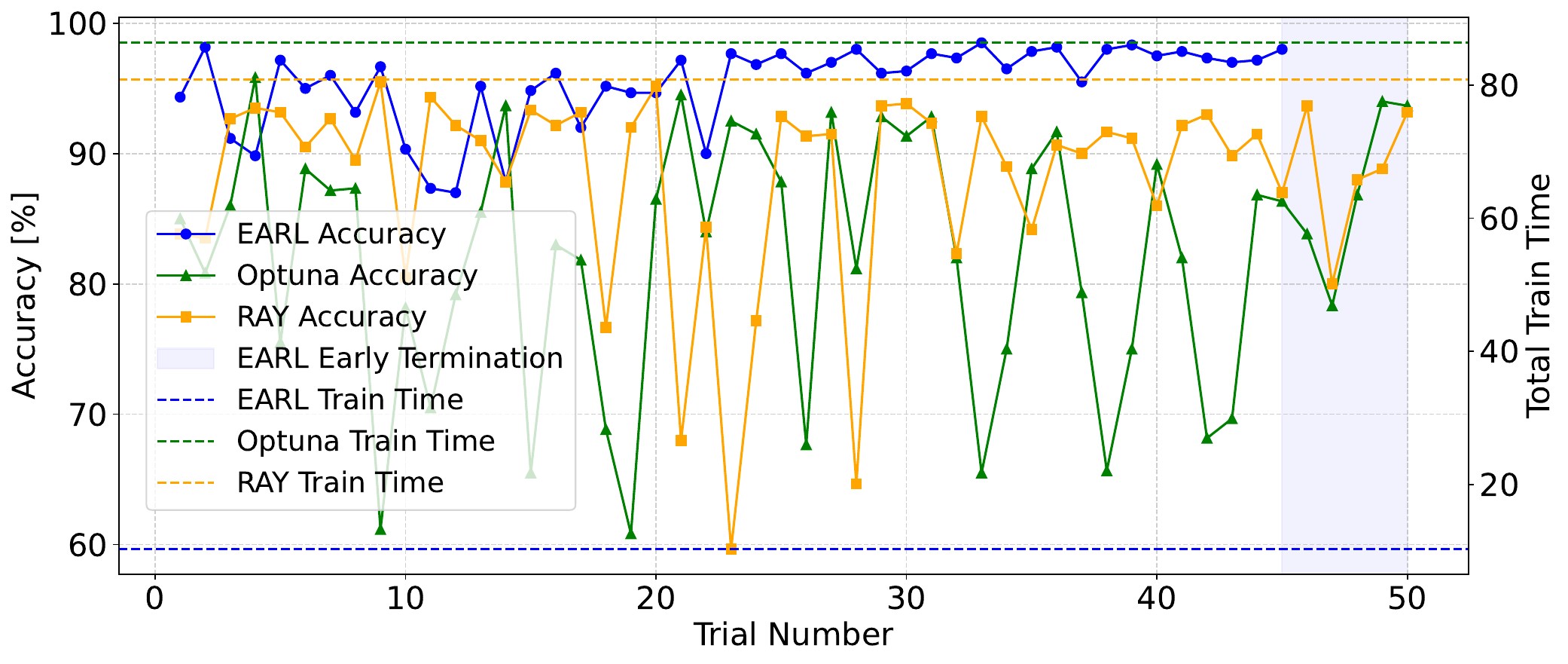}
\label{fig:fsdd-acc}
}
\subfloat[HAR Dataset]{
\includegraphics[width=0.5\columnwidth]{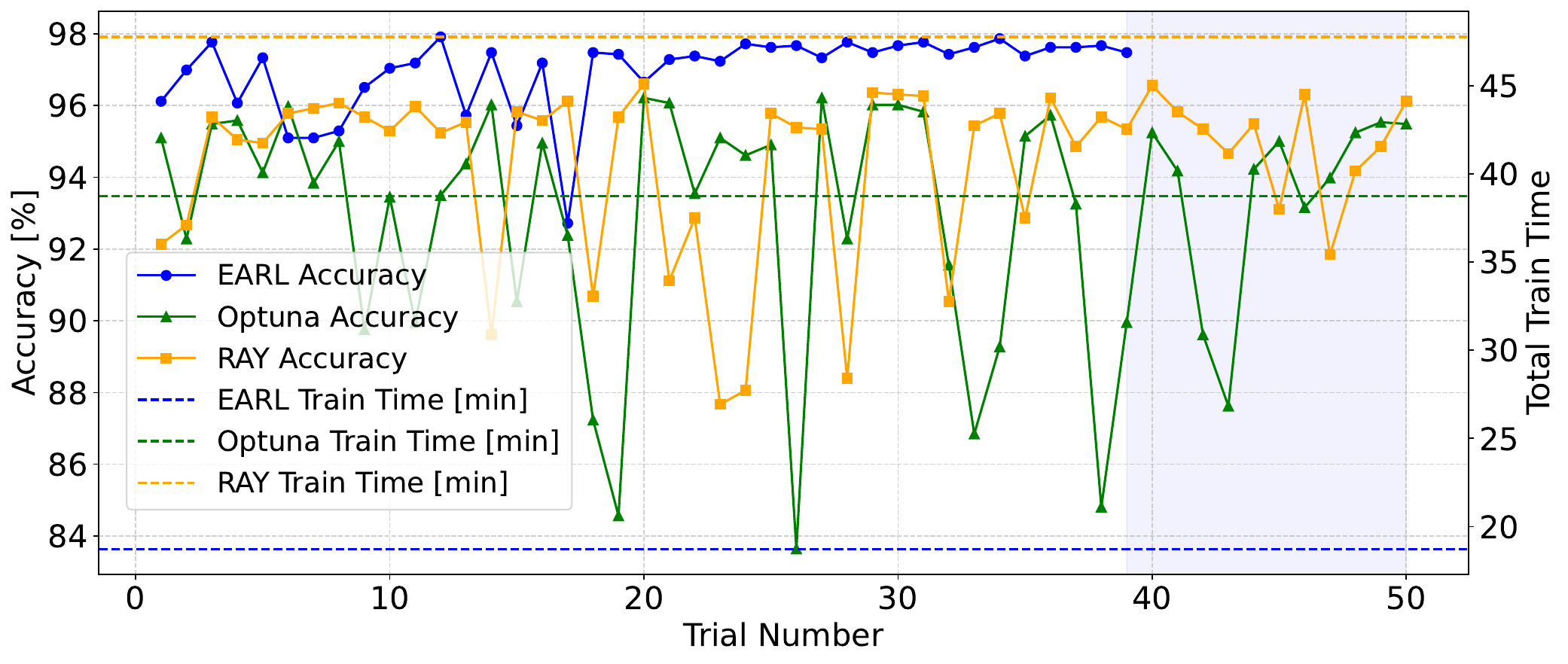}
\label{fig:har-acc}
}\\
\subfloat[Occupancy Dataset]{
\includegraphics[width=0.5\columnwidth]{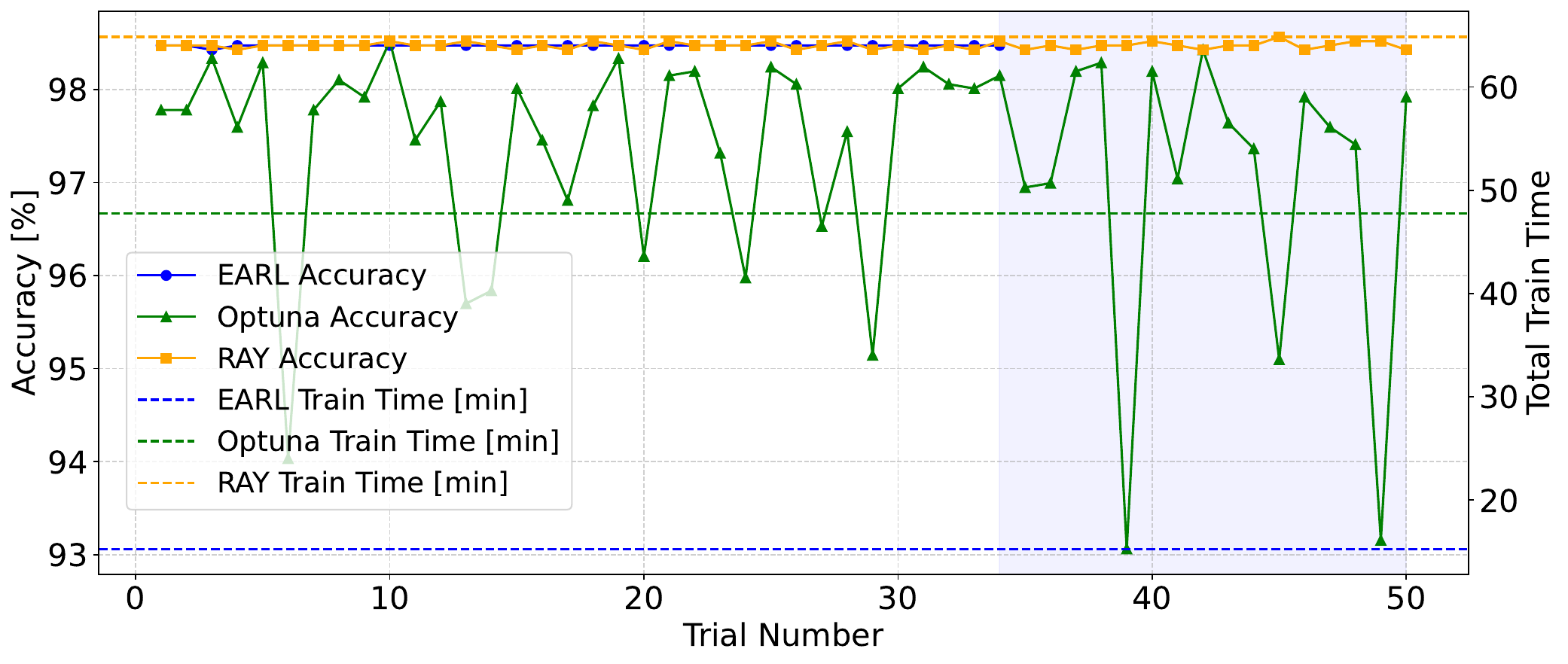}
\label{fig:occ-acc}
}
\caption{Classification accuracy across all optimization models}
\label{fig:accuracy}
\end{figure}

EARL demonstrated superior classification accuracy across all three benchmark datasets, consistently outperforming both Optuna and Ray. On the \textit{FSDD}, EARL reached 95.39\%~$\pm$~0.44, exceeding Optuna (82.25\%~$\pm$~2.75) and Ray (88.15\%~$\pm$~2.18). It stabilized after only 44 trials
(Fig.~\ref{fig:fsdd-acc}) and completed optimization in 10.6 minutes, nearly an order of magnitude faster than both baselines.
On \textit{HAR}, EARL again delivered the best accuracy at 96.99\%~$\pm$~0.34, converging after 39 trials and finishing the search in 19.53 minutes (Fig.~\ref{fig:har-acc}).
For \textit{Occupancy Detection}, EARL attained 
98.47\%~$\pm$~0.00008, matching Ray’s peak accuracy but converging far earlier—after only 34 trials and within 15.28 minutes (Fig~\ref{fig:occ-acc}).

Across all benchmarks, EARL consistently delivered a 6-15\% higher accuracy compared to the baseline methods, with the reinforcement learning component effectively directing the search toward high-potential regions of the hyperparameter space and minimizing unnecessary evaluations.

\subsubsection{Energy Efficiency}
\begin{figure}[htbp!]
\centering
\subfloat[FSDD Dataset]{
\includegraphics[width=0.5\linewidth]{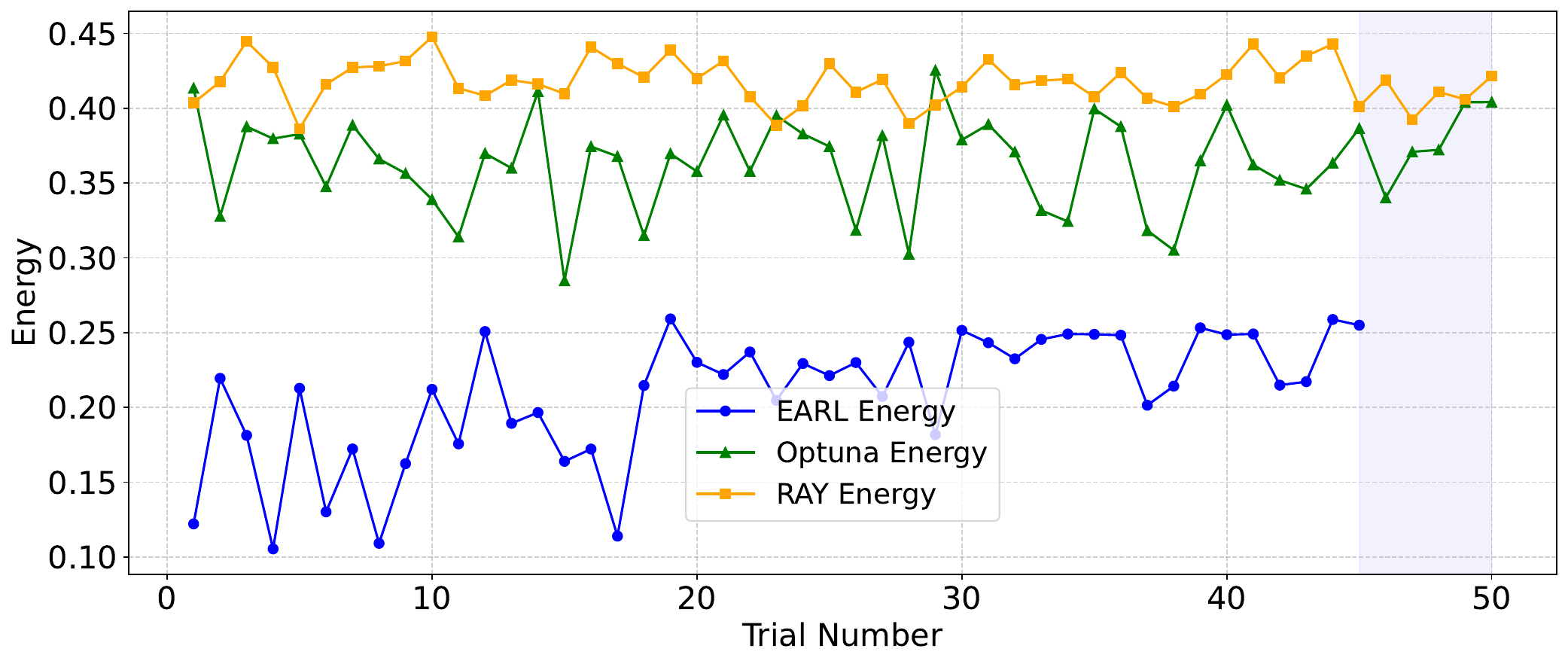}
\label{fig:fsdd-en}
}
\subfloat[HAR Dataset]{
\includegraphics[width=0.5\linewidth]{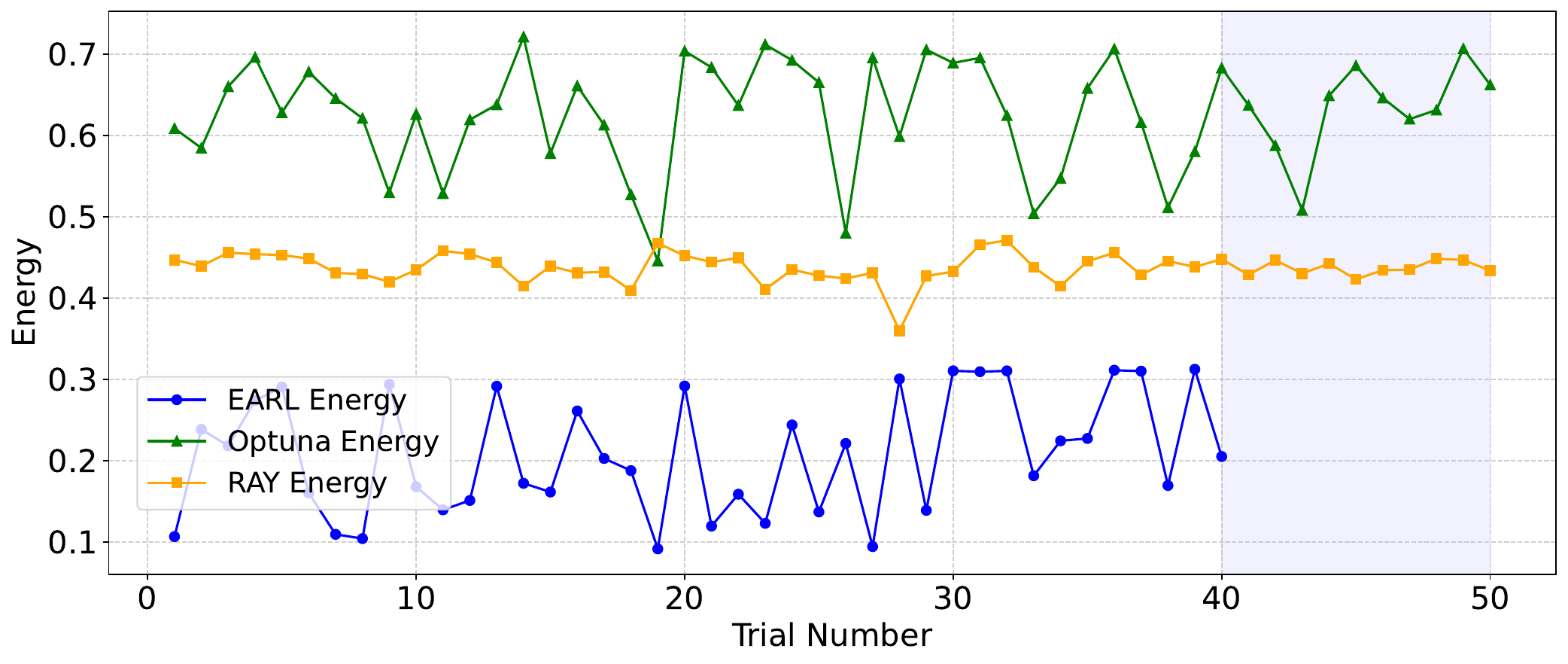}
\label{fig:har-en}
}\\
\subfloat[Occupancy Dataset]{
\includegraphics[width=0.5\linewidth]{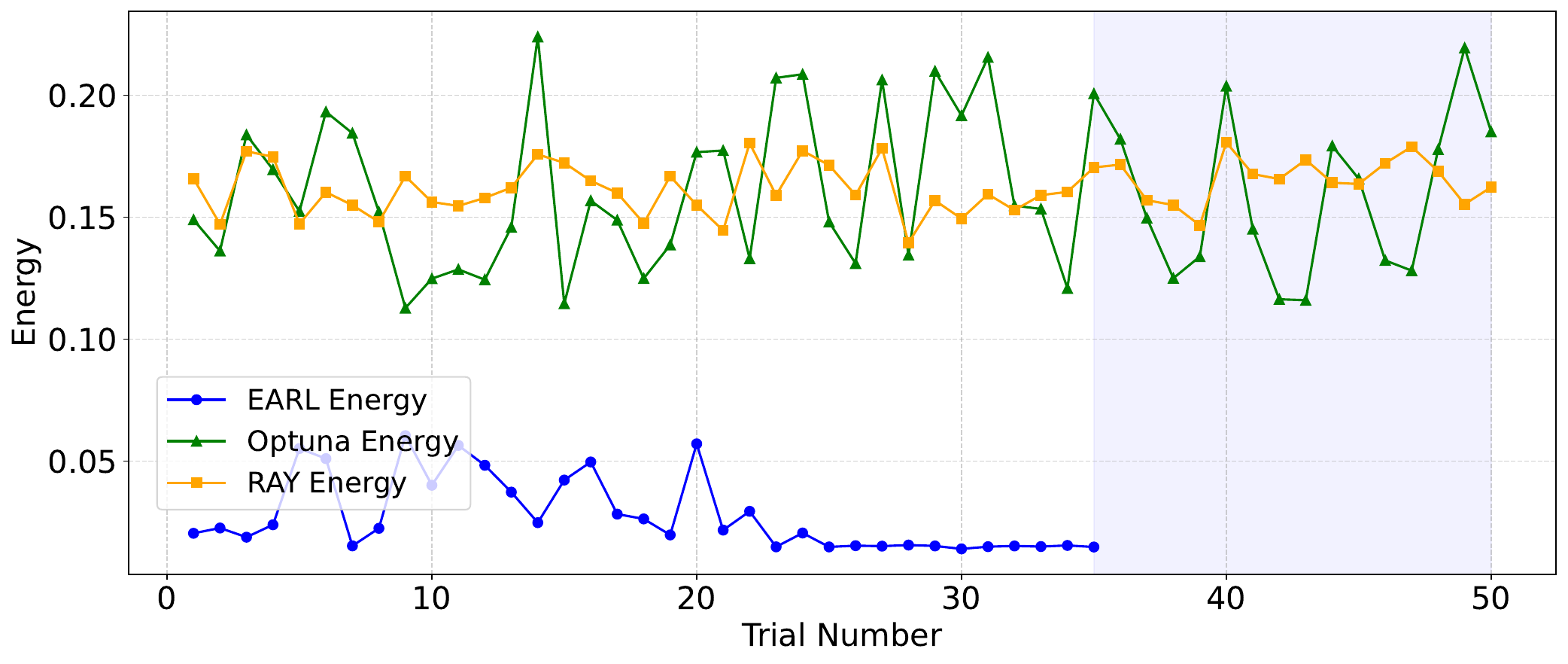}
\label{fig:occ-en}
}
\caption{Energy consumption across all optimization models.}
\label{fig:energy}
\end{figure}

Energy consumption results further highlight EARL's efficiency across all datasets. On FSDD, EARL achieved the lowest mean energy usage at 0.2089~$\pm$~0.0129~pJ per sample, reducing consumption by approximately 50\% relative to the baselines (Fig.~\ref{fig:fsdd-en}). On HAR, EARL maintained its advantage with 0.20796~$\pm$~0.02362~pJ per sample while preserving superior classification accuracy. The largest gains appeared in the Occupancy Detection dataset, where EARL required only 0.0278~$\pm$~0.0052~pJ per sample, compared to Optuna’s
0.1599~$\pm$~0.0092 and Ray’s 0.1624~$\pm$~0.0030~pJ (Figs.~\ref{fig:har-en},\ref{fig:occ-en}). Energy trends remained consistently low throughout the search process.
Overall, EARL reduced energy consumption by 60--80\% across all benchmarks,
demonstrating substantial improvements in computational efficiency.
\subsubsection{Pareto Analysis} 
\begin{figure}[htbp!]
\centering
\subfloat[FSDD Dataset]{
\includegraphics[width=0.5\linewidth]{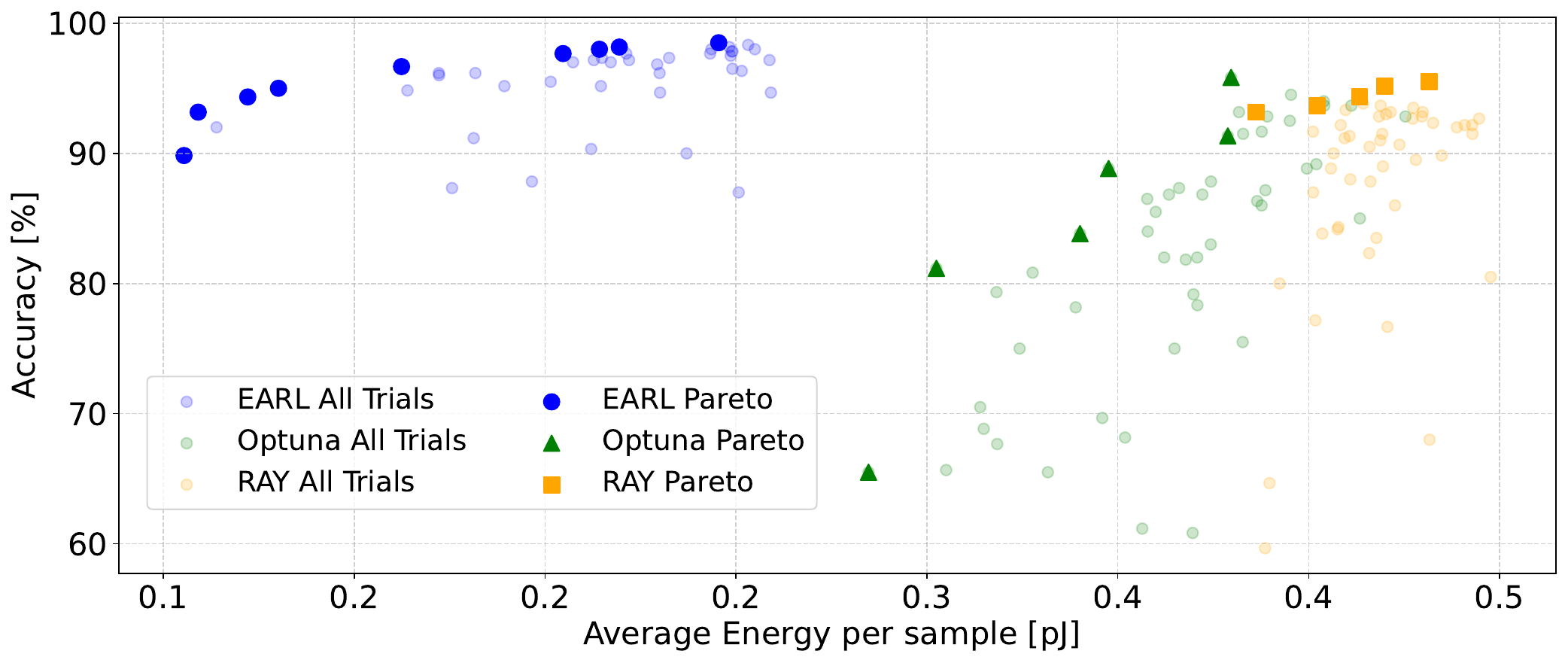}}
\subfloat[HAR Dataset]{
\includegraphics[width=0.5\linewidth]{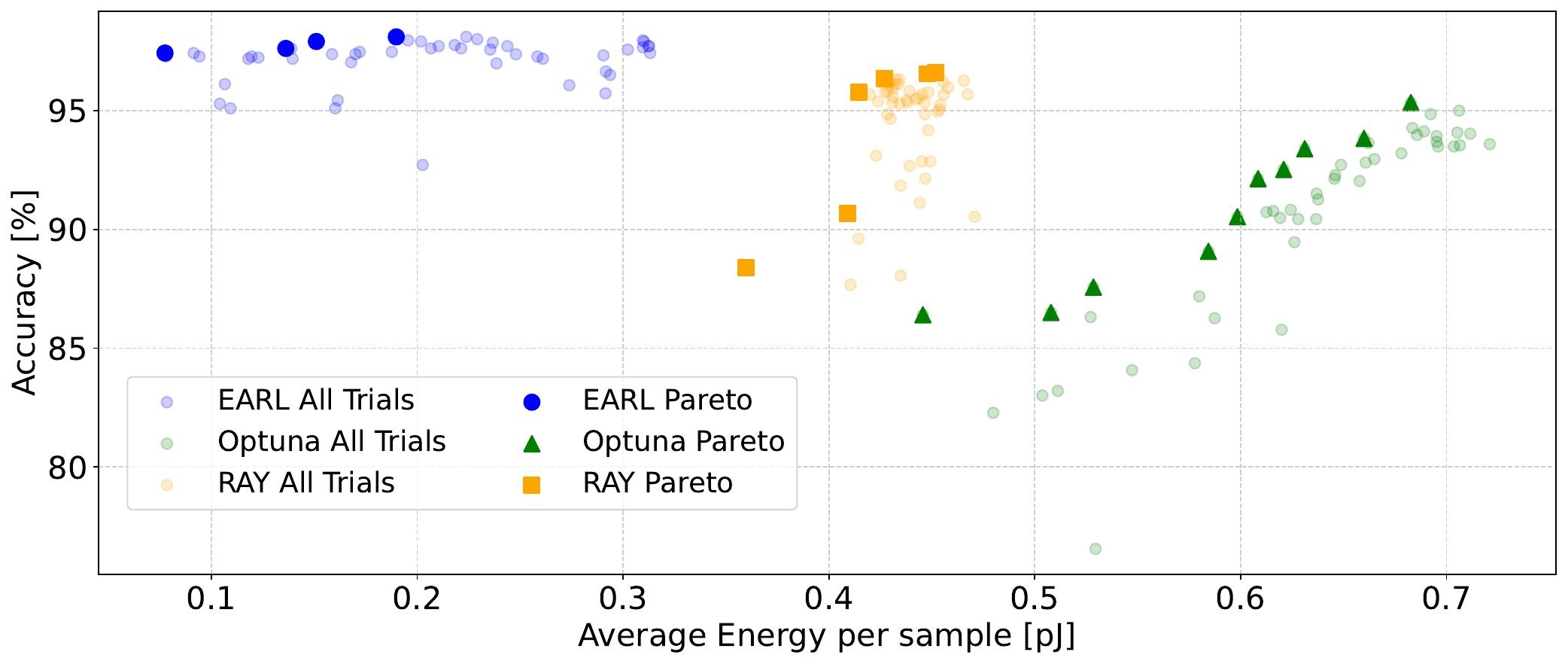}}\\
\subfloat[Occupancy Dataset]{
\includegraphics[width=0.5\linewidth]{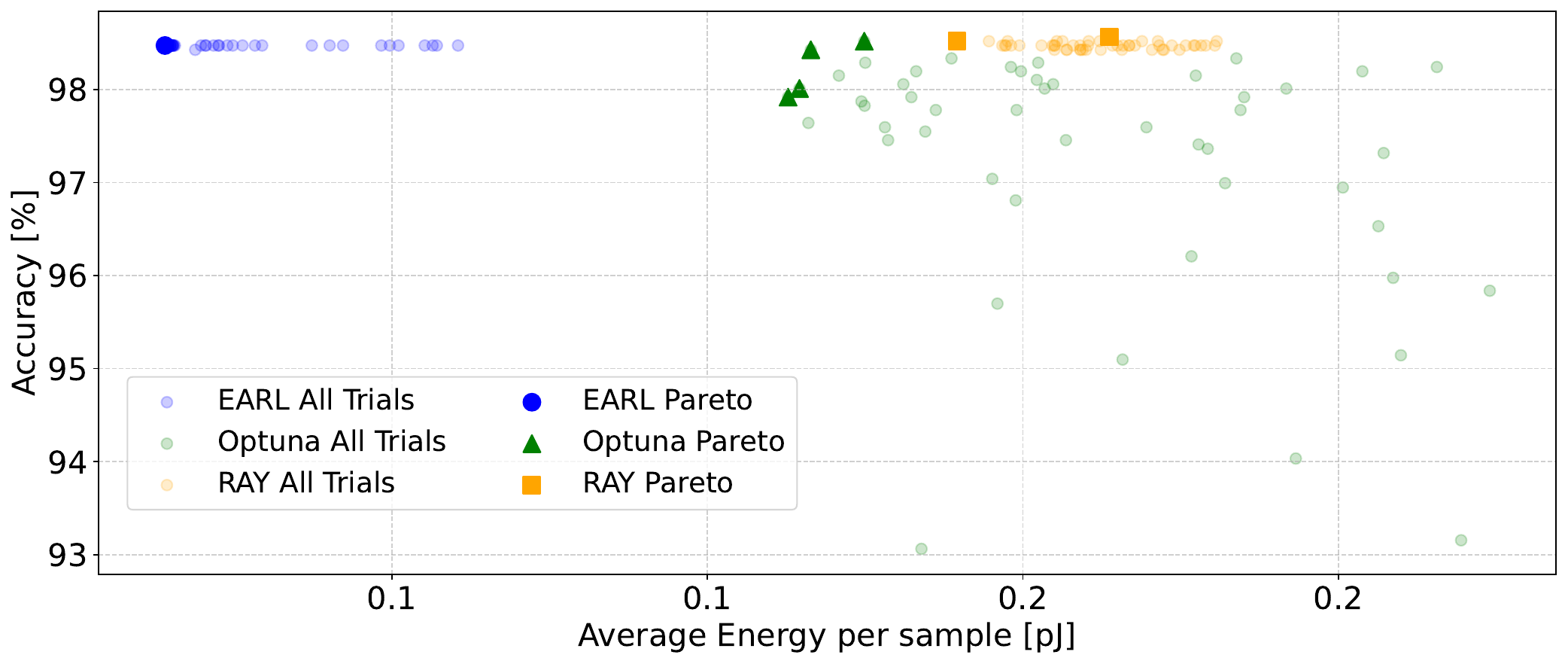}}
\caption{Pareto Front vs All Trials.}
\label{fig:pareto}
\end{figure}

Pareto front analyses across all datasets show that EARL effectively balances accuracy and energy efficiency while identifying optimal configurations. EARL's solutions consistently dominated the high-accuracy, low-energy region with minimal trial-to-trial variability (Fig.~\ref{fig:pareto}). In HAR, the Pareto front reflected a strong accuracy–efficiency trade-off, while in Occupancy Detection EARL clearly produced the most computationally efficient optimal solutions.


\section{Conclusion \& Future Directions}
EARL introduces an energy-aware hyperparameter optimization framework for Liquid State Machines that combines Bayesian optimization with reinforcement learning to prioritize high-accuracy, low-energy configurations. Its adaptive early-termination mechanism further reduces redundant evaluations, yielding substantial computational and energy savings. Across multiple benchmarks, EARL outperforms state-of-the-art baselines such as Optuna and Ray Tune, converging faster and delivering superior accuracy–energy trade-offs. Pareto-front analyses confirm that EARL consistently identifies dominant configurations suited for edge and  deployment. Future work will explore hardware-in-the-loop evaluations on neuromorphic processors and extend the search space to additional architectural parameters to broaden applicability to sensory processing, autonomous systems, and low-power intelligent devices. Overall, EARL represents a scalable, energy-efficient approach to optimizing LSMs for resource-constrained environments.

\bibliographystyle{IEEEtran}
\scriptsize
\bibliography{references}

\end{document}